 \documentclass[final,5p,times,twocolumn,authoryear]{elsarticle}

\usepackage{amssymb}
\usepackage{amsmath}
\usepackage{lipsum}
\usepackage{hyperref}
\usepackage{caption}
\usepackage{multirow}
\usepackage{booktabs} 

\journal{Computer Vision and Image Understanding}

\begin{document}

\begin{frontmatter}

\title{Enhancing Scene Text Detectors with Realistic Text Image Synthesis Using Diffusion Models}

\author[1]{Ling Fu} 
\author[2]{Zijie Wu}
\author[1]{Yingying Zhu\corref{cor1}}
\cortext[cor1]{Corresponding author}
\ead{yyzhu@hust.edu.cn}
\author[2]{Yuliang Liu}

\author[2]{Xiang Bai}

\address[1]{School of Electronic Information
and Communications, Huazhong University of Science and Technology,
Wuhan, 430074, China}
\address[2]{School of Artificial Intelligence and Automation,
Huazhong University of Science and Technology, Wuhan, 430074, China}

\begin{abstract}
Scene text detection techniques have garnered significant attention due to their wide-ranging applications. However, existing methods have a high demand for training data, and obtaining accurate human annotations is labor-intensive and time-consuming. As a solution, researchers have widely adopted synthetic text images as a complementary resource to real text images during pre-training. Yet there is still room for synthetic datasets to enhance the performance of scene text detectors. We contend that one main limitation of existing generation methods is the insufficient integration of foreground text with the background. To alleviate this problem, we present the \textbf{Diff}usion Model based \textbf{Text} Generator (\textbf{DiffText}), a pipeline that utilizes the diffusion model to seamlessly blend foreground text regions with the background's intrinsic features. Additionally, we propose two strategies to generate visually coherent text with fewer spelling errors. With fewer text instances, our produced text images consistently surpass other synthetic data in aiding text detectors. Extensive experiments on detecting horizontal, rotated, curved, and line-level texts demonstrate the effectiveness of DiffText in producing realistic text images. Code will be available at: \href{https://github.com/99Franklin/DiffText}{https://github.com/99Franklin/DiffText}.
\end{abstract}




\end{frontmatter}




\section{Introduction}
\label{sec1}

Scene text detection techniques have seen significant advancements and attracted increased attention from researchers \citep{liu2019scene,long2021scene}. These techniques have many practical applications in areas such as sign localization in the wild \citep{tsai2011mobile}, image retrieval based on scene text \citep{mishra2013image} and assisting visually
impaired users
\citep{chessa2016integrated}. However, existing scene text detection methods require a large quantity of training data. The acquisition of sufficient scene text images and their accurate annotation are labor-intensive and time-consuming processes. To mitigate the above issues, researchers have proposed methods for generating synthetic text images \citep{gupta2016synthetic} and using them as pre-training data. Existing algorithms for generating text images can be roughly categorized into image composition methods and learning-based methods.

\begin{figure}[!t]
\centering
\includegraphics[scale=.18]{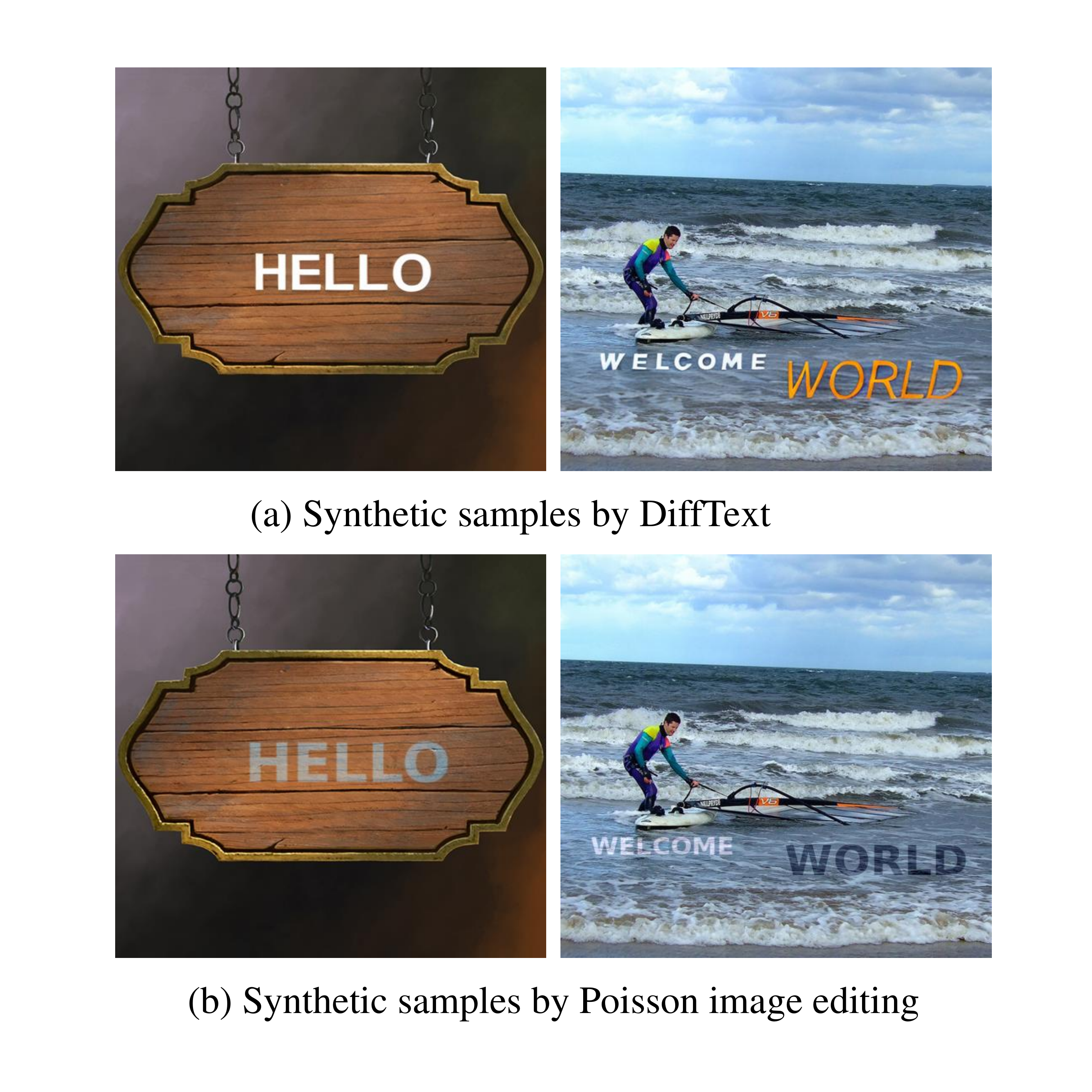}
\caption{Some synthetic samples generated by DiffText and Poisson image editing \citep{perez2003poisson} methods.}
\label{img_first_img}
\end{figure}

Image composition methods \citep{gupta2016synthetic,liao2020synthtext3d, long2020unrealtext,yim2021synthtiger,zhan2018verisimilar} aim to align digitally generated or user-provided text content with background images by optimizing surface smoothness in the combined images. Yet these methods sometimes present noticeable visual discrepancies between the foreground text and the background. These discrepancies mainly stem from challenges in blending different sources. Algorithms such as Poisson image editing \citep{perez2003poisson} used in SynthText \citep{gupta2016synthetic} and the planar meshes employed by UnrealText \citep{long2020unrealtext} mainly rely on visual features in a single image to optimize the blended results.

On the other hand, learning-based methods \citep{tang2022scene,yang2019learning,zhang2019scene} aim to capture the realistic appearance of real text through reference samples. By observing the additional reference data, these methods transform the generated text images to appear visually closer to real text images. Prior works \citep{yang2019learning,zhang2019scene} focused on refining the blended results of text images through mapping learning between unpaired data. Another study \citep{tang2022scene} utilized text segmentation labels to precisely control the appearance learning process from reference images. These approaches mainly rely on learning the visual appearance of reference samples, while the background information from training samples is not fully leveraged.

To seamlessly blend visual text into background images, we leverage the diffusion model as an integration solution. Its robust capability to blend diverse sources makes it a suitable choice for incorporating visual text into scene images. Our pipeline, the Diffusion Model based Text Generator (DiffText), carries out a text-conditional image inpainting task, and can automatically produce realistic text images. Fig. \ref{img_first_img} exhibits some synthetic samples. In the presented cases generated by Poisson image editing, the text contents appear to be pasted on the background image as watermarks. Our method allows the generated text to merge well with backgrounds, presenting superior integration effectiveness. DiffText consists of an autoencoder, a text encoder, and a denoising module. Given a natural image and a selected region for placing the visual text, we first mask the region to obtain a masked image that provides background information. This masked image is then projected into the latent space by the autoencoder, and the denoising module generates the foreground content from Gaussian noise by observing the data distribution of the background, conditioned on the input text string. The fusion of the foreground and background significantly enhances the visual fidelity of the synthesized text compared to previous methods. 

To ensure the quality of the synthetic text images, we employ two strategies during the generation process. Firstly, we crop surrounding regions to obtain local backgrounds, facilitating efficient batch inpainting. Additionally, we utilize a pre-trained text recognizer to filter out low-quality instances. By incorporating these strategies into DiffText, we produced 
high-quality text images, highlighting the potential of our method in generating valuable synthetic data. 
Sufficient ablative experiments demonstrate the advantages of the DiffText. 

We summarize the advantages of our methods as follows:
\begin{itemize} \item To seamlessly integrate visual texts into background images, we present DiffText, a pipeline that uses the diffusion model to blend different sources, enabling the automatic production of realistic text images. \item To enhance the credibility of the generated visual text, we introduce two strategies that are incorporated into DiffText to assist in the generation process. \item We produced 10,000 scene text images by DiffText. With these images, we conducted comprehensive scene text detection experiments on detecting horizontal, rotated, curved, and line-level texts. The results present significant improvements in the performance of scene text detectors.
\end{itemize}

\begin{figure*}[htb]
\centering
\includegraphics[scale=.2]{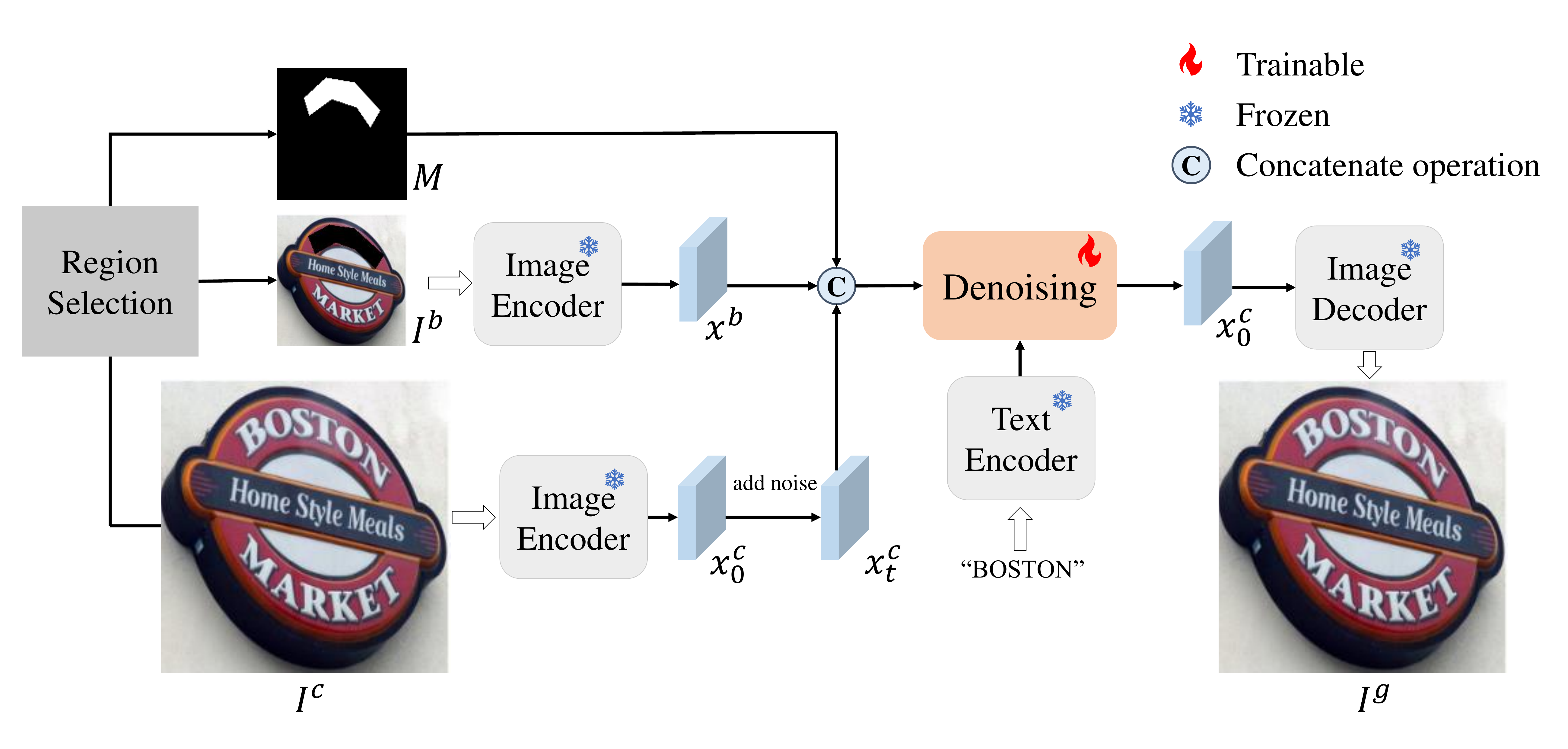}
\caption{The overview framework of our proposed DiffText. Region Selection refers to the process of determining the placement region. Image Encoder and Image Decoder are from the autoencoder. Text Encoder is used to process the input textual string. Denoising refers to the denoising process to predict the added noise.}
\label{img_framwork}
\end{figure*}

\section{Related work}
\label{sec2}
\subsection{Scene text detection}
Scene text detection technology, which aims to automatically localize text within a given image, has been an active research topic. Early text detectors primarily utilized hand-crafted features to search for text regions. For instance, STW \citep{epshtein2010detecting} used edge detection to extract character candidates while MSER \citep{neumann2010method,neumann2012real} relied on extremal region extraction. Additionally, \cite{yi2013text} and \cite{minetto2014snoopertext} respectively utilized character appearances and shape descriptors to help text detectors.
In terms of modern scene text detectors, they can be divided into two categories, i.e., regression-based and segmentation-based methods. 

Regression-based methods explicitly represent the shape of text instances by point sequences. TextBoxes \citep{liao2017textboxes} designed object anchors to fit the shape of text instances. Subsequently, EAST \citep{zhou2017east} adopted pixel-level regression to deal with multi-oriented text instances in an anchor-free approach. To address the issue of Label Confusion, \cite{liu2019omnidirectional,liu2021exploring} proposed to parameterize the bounding boxes into orderless sequences. \cite{zhang2020deep} proposed to regress small rectangular components of text instances and use Graph Convolutional Network to model the connection of these components. To represent the contour of text instances more precisely, Fourier signatures were adopted by FCENet \citep{zhu2021fourier}. To detect texts under extreme traffic scenarios, \cite{he2023domain} proposed to transfer the text detection
ability from conventional scenes to traffic scenes. With the emergence of DETR \citep{carion2020end}, DPText-DETR \citep{ye2023dptext} leveraged learnable queries and attention mechanisms to model the point coordinates within the text region. As for segmentation-based methods, they typically generate the text proposals from pixel-level segmentation maps. \cite{zhang2016multi} used semantic segmentation to predict text regions for multi-oriented instances. To perceive and distinguish adjacent text instances, PSENet \citep{wang2019shape} designed progressive scale expansion and \cite{tian2019learning} introduced pixel embedding. To alleviate the issue of heavy post-processing, DBNet \citep{liao2020real} proposed a differentiable binarization module to perform the binarization process in the network. In our experimental analysis, we selected FCENet and DBNet, both of which excel in balancing performance and efficiency, from each category as text detectors.

\subsection{Text image Generation}
Text Image Generation involves creating visual texts on given natural images and has many applications in text analysis. \cite{wang2012end} proposed a character generator to synthesize training data for character-level text recognizers. To support word-level text recognition models, some methods focused on generating patch images \citep{jaderberg2016reading,fogel2020scrabblegan,kang2020ganwriting,yim2021synthtiger,DBLP:conf/icdar/NikolaidouRCSSSML23, zhu2023conditional}. In addition, other works concentrated on text content generation \citep{xie2021dg, wang2022aesthetic,dai2023disentangling,DBLP:conf/acl/LiuGSCRNBM0C23,ma2023glyphdraw, das2022chirodiff} or text image editing \citep{wu2019editing,shimoda2021rendering,subramanian2021strive,ji2023improving}.

Besides, some studies explore inserting texts into scene images to aim for scene text detectors. Mainstream methods rely on image composition technologies, where digital text foregrounds are overlaid onto background images. Synthtext \citep{gupta2016synthetic} is the pioneering work that proposed to synthesize text images for scene text detectors. Poisson image editing \citep{perez2003poisson} was employed to blend foreground and background sources. Then VISD \citep{zhan2018verisimilar} enhanced the harmonization of text instances using semantic information and a coloring scheme. SynthText3D \citep{liao2020synthtext3d} and UnrealText \citep{long2020unrealtext} leveraged 3D graphics engines to generate text images in 3D scenes. While CurveSynth \citep{liu2021abcnet} aided in detecting arbitrary-shaped texts. Some learning-based methods \citep{tang2022scene,yang2019learning,zhang2019scene} focus on improving the realism of text appearance, but they face challenges due to the limited scope of reference data. In our work, we specifically focus on generating synthetic data that mixes visual texts with the backgrounds for the scene text detection task.

\section{Methodology}
\label{sec3}
The Diffusion Model based Text Generator (DiffText) is a pipeline designed to generate visually appealing text images. DiffText takes a background image and a selected region as input. It then executes a denoising process to transform Gaussian noise into visual content with the given text condition in a latent space. We will describe the details of DiffText and the generation strategies we use to produce synthetic data for scene text detection models in the following subsections.

\begin{figure*}[ht]
\centering
\includegraphics[scale=.15]{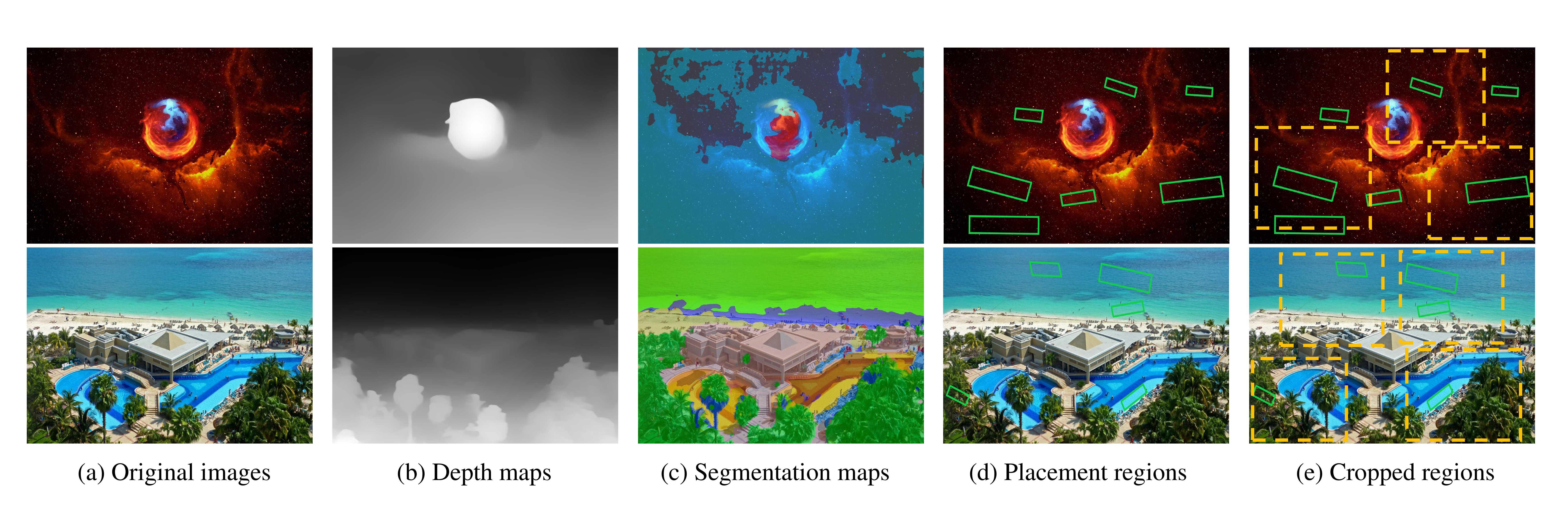}
\caption{(a) The original background images. (b) The corresponding depth maps of the background images. (c) The segmentation maps of the background images. (d) The desired placement regions are indicated by green boxes. (e) The chosen cropped regions are represented by yellow boxes.}
\label{img_depth_seg}
\end{figure*}

\subsection{Synthetic pipeline}

DiffText builds upon Stable Diffusion \citep{rombach2022high}, a method proposed to tackle the computational complexity issue in diffusion models \citep{sohl2015deep}. The diffusion process involves adding noise to the samples, while the denoising process reconstructs the original sample from noised data. These processes are performed at the pixel level. Consequently, when the model's input is a high-resolution image, computational complexity becomes a significant challenge. To this end, Stable Diffusion proposed to encode the image into a latent space with an autoencoder, allowing the diffusion and denoising processes to occur in this space. Additionally, a condition encoder was employed to control the generative model's output under various conditions such as text, semantic maps, and others. More specifically, given an image $x$, the autoencoder translates it into a latent feature $z_0$. Gaussian noise is then added to $z_0$ to yield $z_t$ at the current timestep $t$. Subsequently, a condition encoder $\tau_\theta$ transforms the given condition $y$ into the latent feature $\tau_\theta(y)$. The adopted squared error loss function for predicting noise can be defined as:

\begin{flalign}
\label{eq_1}
    &&
    L = \mathbb{E}_{\epsilon(x),y,\epsilon \sim \mathcal N(0,1),t} \big[\Vert \epsilon - \epsilon_\theta(z_t, t, \tau_\theta(y))\Vert^2_2\big],
    &&
\end{flalign}

Here, $\epsilon$ denotes the actual added noise, while $\epsilon_\theta$ represents the learnable parameters of the denoising module which predicts the added noise.

Our objective is to seamlessly integrate visual text into a natural image while maintaining background consistency to generate realistic text images. In addition to this, we aim to control the text content with the input string. Therefore, a text-conditional image inpainting approach is an ideal solution for these requirements. DiffText consists of three key components: A VAE \citep{esser2021taming} to project image samples from pixel space to the latent space, a text encoder from Clip \citep{radford2021learning} to encode the input textual string. and a denoising module based on UNet \citep{ronneberger2015u} to predict the noise to be added to the latent image feature during the diffusion process. The overview framework of DiffText is illustrated in Fig. \ref{img_framwork}.

To train DiffText, we start by collecting various public scene text image datasets. For each text instance, we mask the corresponding text region using the polygon label. This process creates paired training samples consisting of a masked image $I^b$ and its corresponding original image $I^c$. Initially, $I^c$ is projected into a latent feature $x^c_0$ using the encoder of VAE, denoted as $V_e$. Gaussian noise $\epsilon$ is then iteratively added to the latent feature for $t$ steps to transform it into a noised feature $x^c_t$. The process of obtaining $x^c_t$ can be formulated as:

\begin{flalign}
\label{eq_2}
&&
x^c_t=\sqrt{\overline{\alpha}_t}x^c_0 + \sqrt{1-\overline{\alpha}_t}\epsilon, \quad
\alpha_t = 1 - \beta_t,  \overline{\alpha}_t = \prod_{i=1}^{t} \alpha_{i},
&&
\end{flalign}

where $\beta_i \in [0,1],\forall i \in [1,T]$. $\beta$ is used to schedule the speed of adding noise, which is typically set to increase linearly. Additionally, the background region and mask information are provided to the generation model. To incorporate the additional information, $I^b$ is also projected into the same latent space, resulting in a latent feature $x^b$ that contains the background information. Then $x^c_t$, $x^b$, and a binary map $M$ containing the mask information, are concatenated together. Besides, the desired text string is encoded into a text embedding $t_e$ by the text encoder. Subsequently, the concatenated latent feature $x^m_t$ and $t_e$ are fed into the UNet within the denoising module. The cross-attention \citep{vaswani2017attention} layer is utilized to integrate $t_e$ into the $x^m_t$. Finally, the recovered latent feature is passed through the decoder of VAE, producing the reconstructed image as the output.

During the training process, the parameters of the VAE and the text encoder are kept fixed, while the UNet is trainable. The optimization objective is defined as the squared error loss between the real Gaussian noise and the predicted noise. This can be formulated as:

\begin{flalign}
\label{eq_3}
    &&
    L = \mathbb{E}_{x^m_t,\text{$t_e$},\epsilon \sim \mathcal N(0,1),t} \big[\Vert \epsilon - \epsilon_\theta(x^m_t, t, \text{$t_e$})\Vert^2_2\big],
    &&
\end{flalign}

In Eq. \ref{eq_3}, $\epsilon$ represents the actual added noise, $\epsilon_\theta$ represents the parameters of the UNet model, and $t$ denotes the current timestep for prediction. During the inference phase, the latent feature $x^c_t$ is substituted with Gaussian noise. Similar to the training phase, the masked image and the binary mask are still provided. Then, the denoising module generates the foreground content from the Gaussian noise, using the background information and the given text string. Finally, the decoder of VAE projects the latent feature into the pixel space to generate the text image.

\begin{figure}[!t]
\centering
\includegraphics[scale=.2]{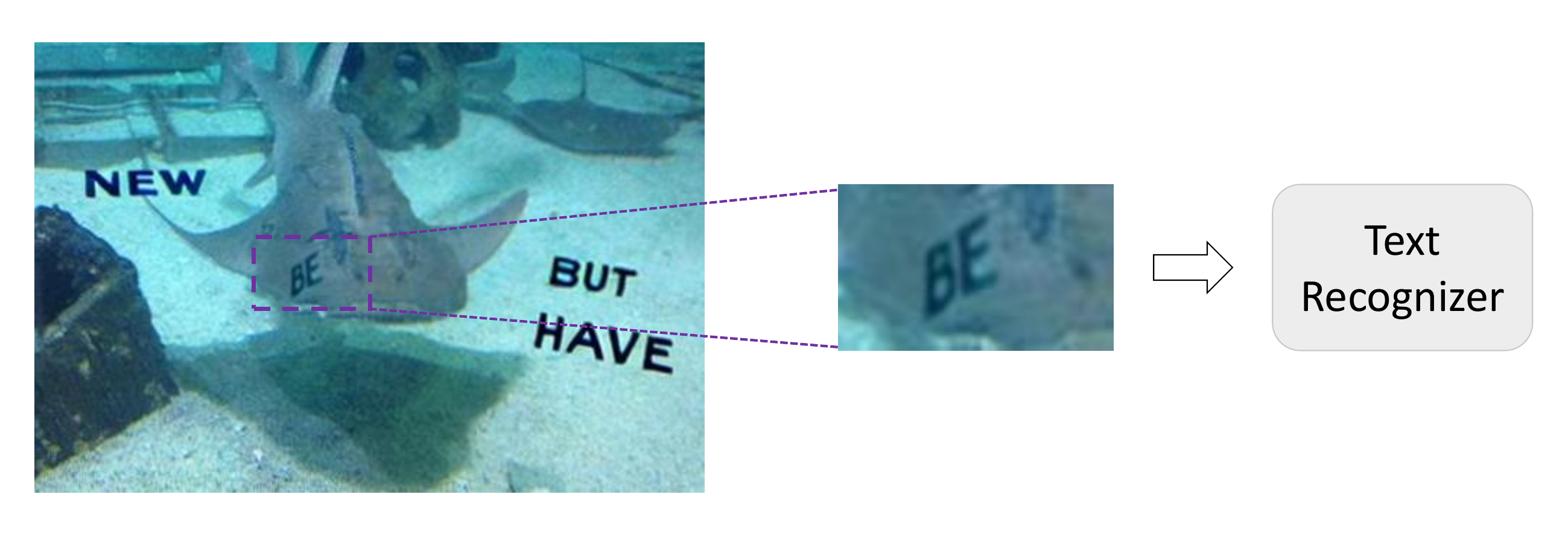}
\caption{Illustration of text instance filtering by a text recognizer. The generated text regions are evaluated by a pretrained text recognizer to determine their preservation.}
\label{img_rec_filter}
\end{figure}

\subsection{Generation strategies}
To generate visually coherent text, we put forth two strategies to aim for the generation process. Firstly, we design a local cropping approach to support forwarding DiffText in batch and reducing the noise during generating. To begin with, we follow the region selection method used in SynthText \citep{gupta2016synthetic}. With the predicted depth information and segmentation maps. Later, we crop local regions from the background image that contain the desired text region(approximately four cropped regions per image). This allows us to forward the diffusion model in batch to expedite faster generation. Furthermore, this cropping operation helps to reduce noise brought to the background during the inpainting process and minimizes interference between the generated text instances.
The entire process is illustrated in Fig. \ref{img_depth_seg}.

Secondly, we propose an instance filtering strategy to help generate credible text. Due to the inherent limitations in character-level content generation \citep{DBLP:conf/acl/LiuGSCRNBM0C23} for the diffusion model, we employ a pretrained text recognizer \citep{fang2021read} to filter out instances with low confidence. Concretely, for the inpainting results of the aforementioned cropped regions, we extract the exact text instance region as a patch image. We then utilize the pretrained text recognizer to process these patch images. If the confidence score falls below a predefined threshold, the instance is discarded. The filtering process is depicted in Fig. \ref{img_rec_filter}. By selecting instances with high confidence scores, we ensure the quality of the generated text. Finally, we replace the cropped regions with the generated regions to finish text rendering.

\section{Experiments}
\label{sec4}

In this section, we provide detailed information about the training process of DiffText and the conducted scene text detection experiments. We begin by introducing the experimental setup, including text image datasets used for comparison, the generation process of our proposed synthetic dataset, as well as the implementation details of DiffText and the text detectors. Following that, we give a comprehensive analysis of the scene text detection experiments.

\subsection{Experimental setup}
\subsubsection{Datasets}
To demonstrate the effectiveness of DiffText, we performed a comprehensive analysis comparing the text images generated by DiffText with previous synthetic datasets, including SynthText \citep{gupta2016synthetic}, VISD \citep{zhan2018verisimilar}, CurveSynth \citep{liu2021abcnet}, SynthText3D \citep{liao2020synthtext3d}, and UnrealText \citep{long2020unrealtext}. Additionally, we used four real text datasets, encompassing ICDAR2013 (IC13) \citep{karatzas2013icdar}, ICDAR2015 (IC15) \citep{karatzas2013icdar}, CTW1500 (CTW) \citep{liu2019curved}, and TotalText \citep{ch2020total}, as test data to evaluate the performance of our approach. In the following paragraphs, we provide a brief introduction to these datasets.

\begin{table}[!t]
\centering
\captionsetup{width=.68\linewidth}
\caption{Calculation of the number of text instances in each synthetic dataset.}\label{tab_text_instances}
\begin{tabular}{p{3cm}p{2cm}}
\hline
Synthetic data & Text instances \\
\hline
SynthText 10K & 84885 \\
VISD 10K & 94359 \\
CurveSynth 10K & 122405 \\
SynthText3D 10K & 143102 \\
UnrealText 10K & 268394 \\
DiffText 10K & 76354 \\
\hline
\end{tabular}
\end{table}

\begin{table*}[!t]
\centering
\captionsetup{width=0.9\linewidth}
\caption{Scene text detection results of DBNet \citep{liao2020real} models trained solely on each synthetic dataset, and tested on each real text dataset without fine-tuning.}\label{tab_pretrain_db}
\begin{tabular}{p{3cm}p{1.05cm}p{1.05cm}p{1.05cm}p{1.05cm}p{1.05cm}p{1.05cm}p{1.05cm}p{1.05cm}p{1.05cm}}
\hline
\multirow{2}{*}{Training data} & \multicolumn{3}{c}{IC13} & \multicolumn{3}{c}{IC15} & \multicolumn{3}{c}{Totaltext} \\
\cmidrule(r){2-4} \cmidrule(r){5-7} \cmidrule(r){8-10}
 & Precision & Recall & Hmean & Precision & Recall & Hmean & Precision & Recall & Hmean \\
\hline
SynthText 10K & 74.77 & 60.64 & 66.97 & 59.02 & 44.25 & 50.58 & 58.61 & 37.34 & 45.62 \\
VISD 10K & 80.15 & 67.85 & 73.49 & 67.12 & 47.57 & 55.68 & 61.88 & 41.85 & 49.93 \\
CurveSynth 10K & 74.29 & 54.89 & 63.13 & 64.13 & 28.41 & 39.37 & 61.61 & 31.74 & 41.90 \\
SynthText3D 10K & 84.62 & 66.85 & 74.69 & 69.53 & 51.85 & 59.40 & 56.30 & 40.72 & 47.26 \\
UnrealText 10K & 81.94 & 65.48 & 72.79 & 67.99 & 43.86 & 53.32 & 48.80 & 29.39 & 36.69 \\
DiffText 10K & 83.88 & 72.24 & \textbf{77.63} & 79.32 & 49.49 & \textbf{60.95} & 65.65 & 44.60 & \textbf{53.12} \\
\hline
\end{tabular}
\end{table*}

\begin{table*}[!t]
\centering
\captionsetup{width=0.9\linewidth}
\caption{Scene text detection results of FCENet \citep{zhu2021fourier} models trained solely on each synthetic dataset, and tested on each real text dataset without fine-tuning.}\label{tab_pretrain_fce}
\begin{tabular}{p{3cm}p{1.05cm}p{1.05cm}p{1.05cm}p{1.05cm}p{1.05cm}p{1.05cm}p{1.05cm}p{1.05cm}p{1.05cm}}
\hline
\multirow{2}{*}{Training data} & \multicolumn{3}{c}{IC13} & \multicolumn{3}{c}{IC15} & \multicolumn{3}{c}{Totaltext} \\
\cmidrule(r){2-4} \cmidrule(r){5-7} \cmidrule(r){8-10}
 & Precision & Recall & Hmean & Precision & Recall & Hmean & Precision & Recall & Hmean \\
\hline
SynthText 10K & 54.59 & 52.69 & 53.62 & 67.67 & 50.79 & 58.03 & 51.84 & 49.53 & 50.66 \\
VISD 10K & 34.08 & 48.58 & 40.06 & 75.15 & 58.98 & 66.09 & 48.68 & 47.54 & 48.10 \\
CurveSynth 10K & 55.80 & 48.31 & 51.79 & 61.37 & 36.78 & 46.00 & 50.47 & 40.99 & 45.24 \\
SynthText3D 10K & 61.84 & 43.65 & 51.18 & 70.06 & 61.29 & 65.38 & 46.33 & 42.98 & 44.59 \\
UnrealText 10K & 68.87 & 50.32 & 58.15 & 74.97 & 57.82 & 65.29 & 54.58 & 44.15 & 48.81 \\
DiffText 10K & 57.96 & 64.47 & \textbf{61.05} & 76.77 & 60.13 & \textbf{67.44} & 57.12 & 55.40 & \textbf{56.25} \\
\hline
\end{tabular}
\end{table*}

\begin{itemize} 
\item SynthText is a widely adopted solution for generating synthetic data to support scene text detection and recognition models. It serves as a valuable resource during the pre-training stage of these algorithms.
\item VISD enhanced the synthetic effect of SynthText by leveraging semantic segmentation information and the proposed color scheme. This results in the generated text appearing more visually harmonious.
\item CurveSynth complements SynthText by providing numerous extra arbitrary-shaped texts. While SynthText mainly focused on generating rotating rectangles, CurveSynth introduced a broader range of text shapes.
\item SynthText3D utilized camera views from Unreal Engine 4, and UnrealCV as the background source. Synthetic digital texts were then rendered onto these scenes to simulate real-world scenarios.
\item UnrealText is another notable work
that employs game engines to generate complex scenes for synthetic text. It utilized collision detection techniques to automatically find suitable placements for the text within the scene.
\item IC13 is a scene text dataset specifically curated for text detection and recognition tasks. In this dataset, the text instances are predominantly horizontally oriented and located at the center of the image.
\item IC15 is another dataset commonly used for text detection and recognition. A notable distinction is that it contains numerous small and blurred text instances, and the text instances are mainly in the shape of rotated rectangles. This poses a significant challenge for the algorithms.
\item CTW contains many line-level instances with arbitrary shapes. It is widely used as a benchmark to evaluate the performance of text detectors, particularly in detecting line-level texts and curved texts.
\item TotalText is another scene text dataset that stands out for containing many arbitrary shape text. This dataset is specifically designed to address the scenario of curved texts, providing valuable resources for the community.

\end{itemize}

\subsubsection{DiffText 10K}
By leveraging the generation capabilities of DiffText, we created 10,000 high-quality text images called DiffText 10K. To ensure a fair comparison with previous synthetic datasets, we randomly selected 10,000 images from SynthText, VISD, CurveSynth, SynthText3D, and UnrealText. These subsets are denoted as SynthText 10K, VISD 10K, CurveSynth 10K, SynthText3D 10K, and UnrealText 10K, and are used in the subsequent comparison experiments. The background images used in our synthetic dataset remain consistent with those from the SynthText dataset, while the text instances are generated using our proposed strategy. The quantities of text instances contained in these synthetic datasets are listed in Tab. \ref{tab_text_instances}. Due to our instance filtering strategy, the number of samples in DiffText 10K is slightly lower than that of SynthText 10K. The generation process was executed on an RTX 3090 GPU and took approximately 30 hours.

\subsubsection{Implementation details}
In all experiments, we utilized PyTorch to implement our models. To train the DiffText, we collected some public scene text datasets, including IC13 \citep{karatzas2013icdar}, IC15 \citep{karatzas2015icdar}, ICDAR2019 \citep{nayef2019icdar2019}, TotalText \citep{ch2020total}, COCO-Text \citep{veit2016coco}, CTW1500 \citep{liu2019curved}, ArT \citep{chng2019icdar2019}, LSVT \citep{sun2019icdar} and TextOCR \citep{singh2021textocr}. The combined total number of text instances in these datasets is 520,337. Each instance is processed by masking the text region and forming a paired sample with the original image for training. To leverage the strong generation capabilities of the Stable Diffusion \citep{rombach2022high}\footnote{https://github.com/huggingface/diffusers}, we initialize the VAE \citep{esser2021taming}, the text encoder of CLIP \citep{radford2021learning}, and the UNet \citep{ronneberger2015u} by loading the parameters of its 2.0-base model. During training, VAE and the text encoder are frozen, while the parameters of UNet are learnable. We utilize the AdamW \citep{DBLP:conf/iclr/LoshchilovH19} optimizer with a learning rate of 1e-5, $\beta1$=0.9, $\beta2$=0.999, and a weight decay of 1e-2. The batch size is set to 24, and the total training epoch is 20. The input images are resized to 512 pixels.

For the scene text detection experiment, we employ two popular text detectors, a segmentation-based text detector (DBNet) \citep{liao2020real} and a regression-based text detector (FCENet) \citep{zhu2021fourier}. The training strategies follow the default configuration of mmocr's implementation\footnote{https://github.com/open-mmlab/mmocr}. Specifically, for DBNet, the pretraining procedure adopts Stochastic gradient descent (SGD) as the optimizer with a learning rate of 0.007, momentum of 0.9, and a weight decay of 1e-4 for training 100,000 iterations. During the fine-tuning procedure, the model is trained for 1200 epochs. For FCENet, in the pretraining procedure, the learning rate is set to 0.001, weight decay is set to 5e-4, and the other settings remain the same as those in DBNet. All experiments are conducted on RTX 3090 GPUs.

\begin{table}[!t]
\centering
\captionsetup{width=0.98\linewidth}
\caption{Scene text detection results of DBNet \citep{liao2020real} models trained on each synthetic dataset and IC15 \citep{karatzas2015icdar}. The DBNet models are initially trained solely on each synthetic dataset, followed by fine-tuning on IC15, and then tested on IC15. $\dag$ denotes that the DBNet models utilize the parameters of ResNet-50 \citep{he2016deep} trained on ImageNet \citep{deng2009imagenet}.}\label{tab_finetune}
\begin{tabular}{p{3.7cm}p{1.05cm}p{1.05cm}p{1.05cm}}
\hline
\multirow{2}{*}{Training data} & \multicolumn{3}{c}{IC15} \\
\cmidrule(r){2-4}
 & Precision & Recall & Hmean \\
\hline
IC15 & 87.10 & 75.73 & 81.02 \\
$\text{IC15}^\dag$ & 87.20 & 81.66 & 84.34 \\
IC15 + SynthText 10K & 88.38 & 83.10 & 85.66 \\
IC15 + VISD 10K & 88.68 & 82.57 & 85.51 \\
IC15 + CurveSynth 10K & 89.33 & 82.62 & 85.84 \\
IC15 + SynthText3D 10K & 87.72 & 82.52 & 85.04 \\
IC15 + UnrealText 10K & 88.76 & 82.86 & 85.71 \\
IC15 + DiffText 10K & 89.79 & 83.44 & \textbf{86.50} \\
\hline
\end{tabular}
\end{table}

\begin{table}[!t]
\centering
\captionsetup{width=0.98\linewidth}
\caption{Scene text detection results of DBNet \citep{liao2020real} models trained on each synthetic dataset and CTW \citep{liu2019curved} (short for SCUT-CTW1500). The DBNet models are initially trained solely on each synthetic dataset, followed by fine-tuning on CTW, and then tested on CTW. $\dag$ denotes that the DBNet models utilize the parameters of ResNet-50 \citep{he2016deep} trained on ImageNet \citep{deng2009imagenet}.}\label{tab_finetune_ctw}
\begin{tabular}{p{3.7cm}p{1.05cm}p{1.05cm}p{1.05cm}}
\hline
\multirow{2}{*}{Training data} & \multicolumn{3}{c}{CTW} \\
\cmidrule(r){2-4}
 & Precision & Recall & Hmean \\
\hline
CTW & 71.25 & 65.29 & 68.14 \\
$\text{CTW}^\dag$ & 72.00 & 69.27 & 70.61 \\
CTW + SynthText 10K & 75.79 & 68.01 & 71.69 \\
CTW + VISD 10K & 74.67 & 68.01 & 71.18 \\
CTW + CurveSynth 10K & 74.66 & 69.94 & 72.22 \\
CTW + SynthText3D 10K & 74.32 & 68.90 & 71.51 \\
CTW + UnrealText 10K & 74.49 & 69.20 & 71.75 \\
CTW + DiffText 10K & 76.31 & 71.06 & \textbf{73.59} \\
\hline
\end{tabular}
\end{table}

\begin{table*}[!t]
\centering
\captionsetup{width=0.91\linewidth}
\caption{Scene text detection results of DBNet \citep{liao2020real} trained on different synthetic data under various setting. ``With crop'' refers to the generation process with region crop strategy, ``With rec'' indicates the generation process with instances filtering strategy.}\label{tab_ablation}
\begin{tabular}{p{1.3cm}p{1.3cm}p{1.05cm}p{1.05cm}p{1.05cm}p{1.05cm}p{1.05cm}p{1.05cm}p{1.05cm}p{1.05cm}p{1.05cm}}
\hline
\multirow{2}{*}{With crop} & \multirow{2}{*}{With rec} & \multicolumn{3}{c}{IC13} & \multicolumn{3}{c}{IC15} & \multicolumn{3}{c}{Totaltext} \\
\cmidrule(r){3-5} \cmidrule(r){6-8} \cmidrule(r){9-11}
 & & Precision & Recall & Hmean & Precision & Recall & Hmean & Precision & Recall & Hmean \\
\hline
 &  & 81.97 & 61.46 & 70.25 & 59.28 & 39.53 & 47.43 & 62.32 & 41.44 & 49.78 \\
\resizebox{12pt}{!}{\checkmark} &  & 78.48 & 64.29 & 70.68 & 69.19 & 41.41 & 51.81 & 59.99 & 42.57 & 49.80 \\
 & \resizebox{12pt}{!}{\checkmark} & 82.23 & 65.94 & 73.19 & 67.87 & 39.96 & 50.30 & 67.09 & 43.34 & 52.66 \\
\resizebox{12pt}{!}{\checkmark} & \resizebox{12pt}{!}{\checkmark} & 83.88 & 72.24 & \textbf{77.63} & 79.32 & 49.49 & \textbf{60.95} & 65.65 & 44.60 & \textbf{53.12} \\
\hline
\end{tabular}
\end{table*}

\begin{figure*}[!t]
\centering
\includegraphics[scale=.16]{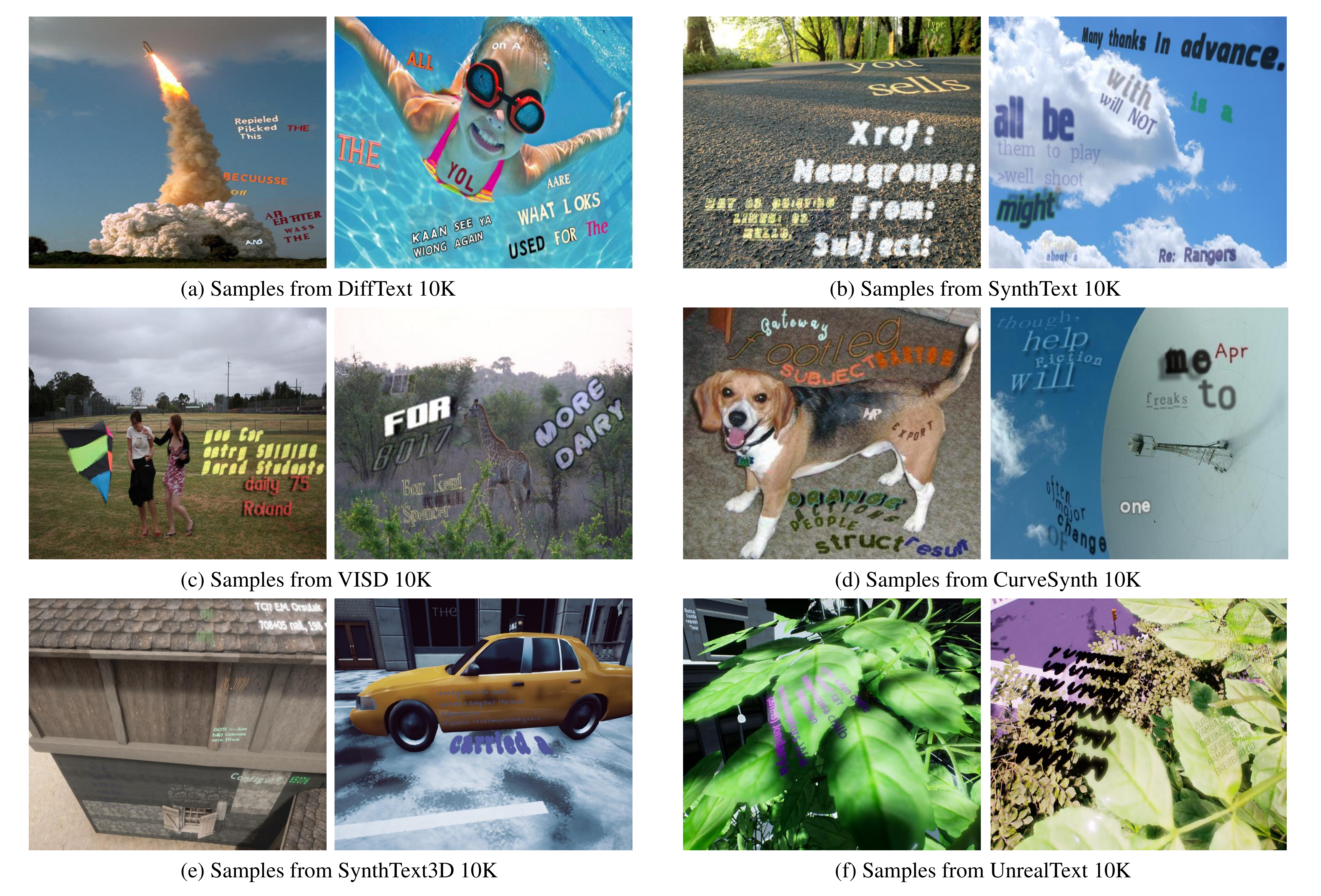}
\caption{Some samples from each synthetic dataset. (a) Samples from the DiffText; (b) samples from the SynthText \citep{gupta2016synthetic}; (c) samples from VISD \citep{zhan2018verisimilar}; (d) samples from CurveSynth \citep{liu2021abcnet}; (e) samples from SynthText3D \citep{liao2020synthtext3d}; (f) samples from UnrealText \citep{long2020unrealtext}.}
\label{img_demo}
\end{figure*}

\subsection{Comparison with previous methods}

\textbf{Only synthetic data.} We trained two popular text detectors, DBNet \citep{liao2020real} and FCENet \citep{zhu2021fourier}, with each synthetic text dataset to demonstrate the effectiveness of DiffText. The experiment results are presented in Tab. \ref{tab_pretrain_db} and Tab. \ref{tab_pretrain_fce}. From the tables, we observe that the text detectors trained on synthetic data perform well on real text datasets. This demonstrates the necessity of generating synthetic data, especially when high-quality text annotations are scarce. Furthermore, the text detector trained on DiffText 10K consistently outperforms other text detectors on all datasets, validating the realism of text in DiffText 10K and the effective density estimation capability of DiffText.

We present samples from DiffText 10K along with other synthetic datasets in Fig. \ref{img_demo}. Due to the limitations in blending foreground texts and background, the scene texts in SynthText \citep{gupta2016synthetic}, VISD \citep{zhan2018verisimilar} and CurveSynth \citep{liu2021abcnet} visually appear to be pasted onto the background images. On the other hand, SynthText3D \citep{liao2020synthtext3d} and UnrealText \citep{long2020unrealtext} exhibit better harmony between the texts and the backgrounds. This is mainly because both the texts and the backgrounds in these datasets are digitally simulated, resulting in a more cohesive visual appearance. However, there still exists an inevitable domain gap between the simulated scenes and real images. In contrast, DiffText overcomes these limitations by learning the data distribution of entire real text images and seamlessly integrating text regions into the background. This approach ensures that our generated texts appear more realistic and provide greater benefits to text detectors. It is noteworthy that the number of text instances in DiffText 10K is significantly lower compared to other methods, as described in Tab.\ref{tab_text_instances}, which supports our claim.

\textbf{Synthetic data and real data.} To further demonstrate the advantages of our generated text images, we conducted experiments by training text detectors using a combination of real and synthetic data. Initially, the DBNet \citep{liao2020real} models were trained on synthetic data, followed by fine-tuning on real text images. The results of these experiments are presented in Tab. \ref{tab_finetune}. Through pretraining on synthetic data, the text detectors significantly improved their performance on the test datasets. The gains achieved by incorporating 10K synthetic data into the text detectors surpassed the improvements obtained from well-trained parameters sourced from ImageNet \citep{deng2009imagenet}. This highlights the necessity of utilizing synthetic data for pre-training. Notably, due to the realistic nature of the text in DiffText 10K, the corresponding text detectors inherited enhanced visual text perception abilities during pre-training. Consequently, DiffText 10K provided better benefits for the text detectors compared to other synthetic text datasets.

Moreover, we replace the real data of the previous experiment from IC15 to CTW1500 and replicate the fine-tuning process to evaluate the performance on line-level texts and curved texts. The results are displayed in Tab. \ref{tab_finetune_ctw}. Despite the lack of curved text instances, DiffText 10K still presents superior performance. This indicates that our text images can effectively improve the robust generalization ability of the text detector.

\subsection{Ablation studies}

To evaluate the effectiveness of our proposed generation strategies, we conducted ablation studies. Synthetic data was generated under various settings and used as training data for the DBNet \citep{liao2020real} text detectors. The results on each real text dataset are presented in Tab. \ref{tab_ablation}. ``With Crop'' refers to the generation process utilizing the region crop strategy, while ``With Rec'' indicates the generation process involving instances filtering strategy with text recognizer. 

From the table, it is evident that the detection performance significantly drops without the region crop strategy or instances filtering strategy. As mentioned earlier, the region crop strategy helps the model reduce noise introduced by the inpainting process and diminish interference among generated text instances. On the other hand, the instances filtering strategy mitigates the issue of generating unstable text, thereby enhancing the production of discriminative text instances.

\section{Discussion}

Our studies establish the necessity of synthetic data for training scene text detectors. Furthermore, our produced high-quality text images yield greater benefits. Compared to previous methods, DiffText enhances the realism of visual text by seamlessly integrating the foreground content with the background. In future studies, more advanced generation models will produce more legible text content, thereby improving the performance of scene text spotters and helping many other downstream applications. Additionally, the efficiency of the generation process could be further optimized to expedite the production of large amounts of text images, which can support the further advancement of scene text detection technologies.

\section{Conclusion}

In this paper, we present DiffText, a pipeline that seamlessly integrates foreground text into background images to generate realistic scene text images. DiffText treats text image generation as a text-conditional image inpainting task. Specifically, it fulfills the foreground content by observing the intrinsic features of the background region and the text condition. Using DiffText, we produced a collection of high-quality text images, highlighting the potential of DiffText in generating valuable text datasets. Furthermore, we conducted comprehensive scene text detection experiments to demonstrate the realism of these text images. Experiment results show that our generated text images exhibit a diminished domain gap compared to previous synthetic data, resulting in significant benefits for scene text detectors. 





\bibliographystyle{elsarticle-harv} 
\bibliography{main}






\end{document}